\documentclass[letterpaper]{article} 
\usepackage{aaai2026}  
\usepackage{times}  
\usepackage{helvet}  
\usepackage{courier}  
\usepackage[hyphens]{url}  
\usepackage{graphicx} 
\usepackage{natbib}  
\usepackage{caption} 
\usepackage{siunitx}
\usepackage{amsmath}
\usepackage{amssymb}
\usepackage{multirow}
\usepackage{adjustbox}
\usepackage{booktabs}
\usepackage{comment}
\usepackage{makecell}
\usepackage{float}
\usepackage{multicol}
\usepackage{hhline}
\usepackage{array}
\usepackage{pgfplots}
\usepackage{tcolorbox}
\usepackage{graphicx}
\usepackage{subcaption}
\usepackage{xspace}
\usepackage{newfloat}
\usepackage{listings}
\DeclareCaptionStyle{ruled}{labelfont=normalfont,labelsep=colon,strut=off} 
\floatstyle{ruled}
\newfloat{listing}{tb}{lst}{}
\floatname{listing}{Listing}
\newcommand{\ourmodel}{\textsc{Smart}\xspace}

\title{\ourmodel: A Surrogate Model for Predicting Application Runtime in Dragonfly Systems}

\author{
    Xin Wang\equalcontrib,\textsuperscript{\rm 1}
    Pietro Lodi Rizzini\equalcontrib,\textsuperscript{\rm 1}
    Sourav Medya,\textsuperscript{\rm 1}
    Zhiling Lan\textsuperscript{\rm 1,2}
}
\affiliations{
    \textsuperscript{\rm 1}University of Illinois Chicago\\
    \textsuperscript{\rm 2}Argonne National Laboratory\\
    \{xwang823, plodir2, medya, zlan\}@uic.edu
    
}

\begin{document}

\maketitle

\begin{abstract}
The Dragonfly network, with its high-radix and low-diameter structure, is a leading interconnect in high-performance computing. A major challenge is workload interference on shared network links.
Parallel discrete event simulation (PDES) is commonly used to analyze  workload interference. However, high-fidelity PDES is computationally expensive, making it impractical for large-scale or real-time scenarios. Hybrid simulation that incorporates data-driven surrogate models offers a promising alternative, especially for forecasting application runtime, a task complicated by the dynamic behavior of network traffic.
We present \ourmodel, a surrogate model that combines graph neural networks (GNNs) and large language models (LLMs) to capture both spatial and temporal patterns from port level router data. \ourmodel outperforms existing statistical and machine learning baselines, enabling accurate runtime prediction and supporting efficient hybrid simulation of Dragonfly networks.
\end{abstract}

\begin{links}
    \link{Code}{https://github.com/SPEAR-UIC/SMART}
    \link{Datasets}{https://zenodo.org/records/16667461}
\end{links}

\section{Introduction}
\par High-performance computing (HPC) systems play a vital role in numerous scientific domains, including climate modeling and drug discovery \cite{hpcClimate}, where complex simulations and data-intensive computations are required. In these systems, the interconnect network serves as the central nervous system, facilitating data exchange among computer nodes and hence determining the system's overall performance. Among the various network topologies designed to meet these demands, the Dragonfly topology stands out as a high-radix, low-diameter solution that balances cost and performance \cite{dragonfly}. Distinct from fat-tree networks, Dragonfly networks reduce infrastructure costs while maintaining high throughput and scalability. This effectiveness is reflected in their widespread adoption in the fast supercomputers, as evidenced by the latest Top500 list (6/2025), where six of the top ten supercomputers utilize Dragonfly interconnects \cite{top500}.

\par In Dragonfly systems, a critical challenge is workload interference, which arises as multiple applications run concurrently on the same machine, producing highly dynamic and complex network traffic patterns. Parallel discrete event simulation (PDES) is widely used to model workload interference and network flows with high fidelity, i.e., at the flit-level granularity \cite{pdes}.
CODES is a widely recognized PDES-based network simulator known for its ability to accurately model Dragonfly network behavior in detail\cite{mubarak2017enabling,ross,union}. However, the computational complexity of high-fidelity PDES grows intractably, requiring the processing of billions of flit-level events across thousands of routers. 
As a result,  hybrid simulation integrating data-driven surrogate models into PDES has become increasingly important for accelerating network modeling while preserving accuracy. 
One particular area of interest is constructing surrogate models to predict application runtime in Dragonfly systems.

\par Developing \emph{data-driven surrogate models} for PDES is extremely challenging due to the highly dynamic nature of network traffic (e.g., at the millisecond scale) and the complex performance fluctuations inherent in large-scale systems. These challenges are exacerbated when multiple parallel applications run concurrently, competing for shared network resources, with iteration times that shift dynamically in response to changing network loads. Additionally, surrogate models must strike a delicate balance: they must achieve high accuracy while imposing minimal runtime overhead to be practical for real-time use. Such surrogate models hold significant potential to enhance networking systems by enabling improved decision-making in areas such as routing, congestion management, and resource allocation.

In this work, we explore advanced machine learning for surrogate modeling of Dragonfly interconnects, aiming to advance PDES and, in turn, contribute meaningful advancements to Dragonfly network research.
The major contributions are summarized as follows. 

\vspace{-0.1cm}
\begin{itemize}
    \item \textbf{Objective:}  Our goal is to develop an accurate yet lightweight surrogate model for predicting application iteration times in Dragonfly. Such a model can not only accelerate networking simulations, but also contributes to improving networking decisions in Dragonfly systems.
\item \textbf{Novel Approach:} We develop \textbf{\ourmodel} (\underline{\textbf{S}}urrogate \underline{\textbf{M}}odel for Predicting \underline{\textbf{A}}pplication \underline{\textbf{R}}un\underline{\textbf{T}}ime).
Our design leverages domain-specific knowledge of Dragonfly interconnects by incorporating both \emph{topological structure} and \emph{temporal dynamics} of network traffic to address the application runtime forecasting problem.
Unlike the conventional algorithms, {\ourmodel} integrates graph neural networks (GNNs) and large language models (LLMs) to holistically model the complex behavior of Dragonfly systems. 
Specifically, GNNs are employed to capture the hierarchical and spatial connectivity inherent in the Dragonfly topology, while LLMs are introduced to enhance temporal modeling through long-range pattern recognition and contextual prompting, a capability unavailable in traditional sequence models. To the best of our knowledge, this is the first effort of building a surrogate model for large-scale networking using LLMs and GNNs. 
    \looseness=-1
\item \textbf{Experimental Results:} We conduct extensive experiments using two datasets generated from a 1,056-node Dragonfly system, where multiple representative HPC workloads are executed concurrently under various job placement strategies. Our method {\ourmodel} significantly outperforms all the baselines in predictive accuracy. Moreover, {\ourmodel} achieves an inference time of just 0.515 seconds, offering at least an order-of-magnitude speedup over the original simulation time.
\end{itemize}

\section{Background}
\begin{figure}[t]
    \centering
    \includegraphics[width=0.85\columnwidth]{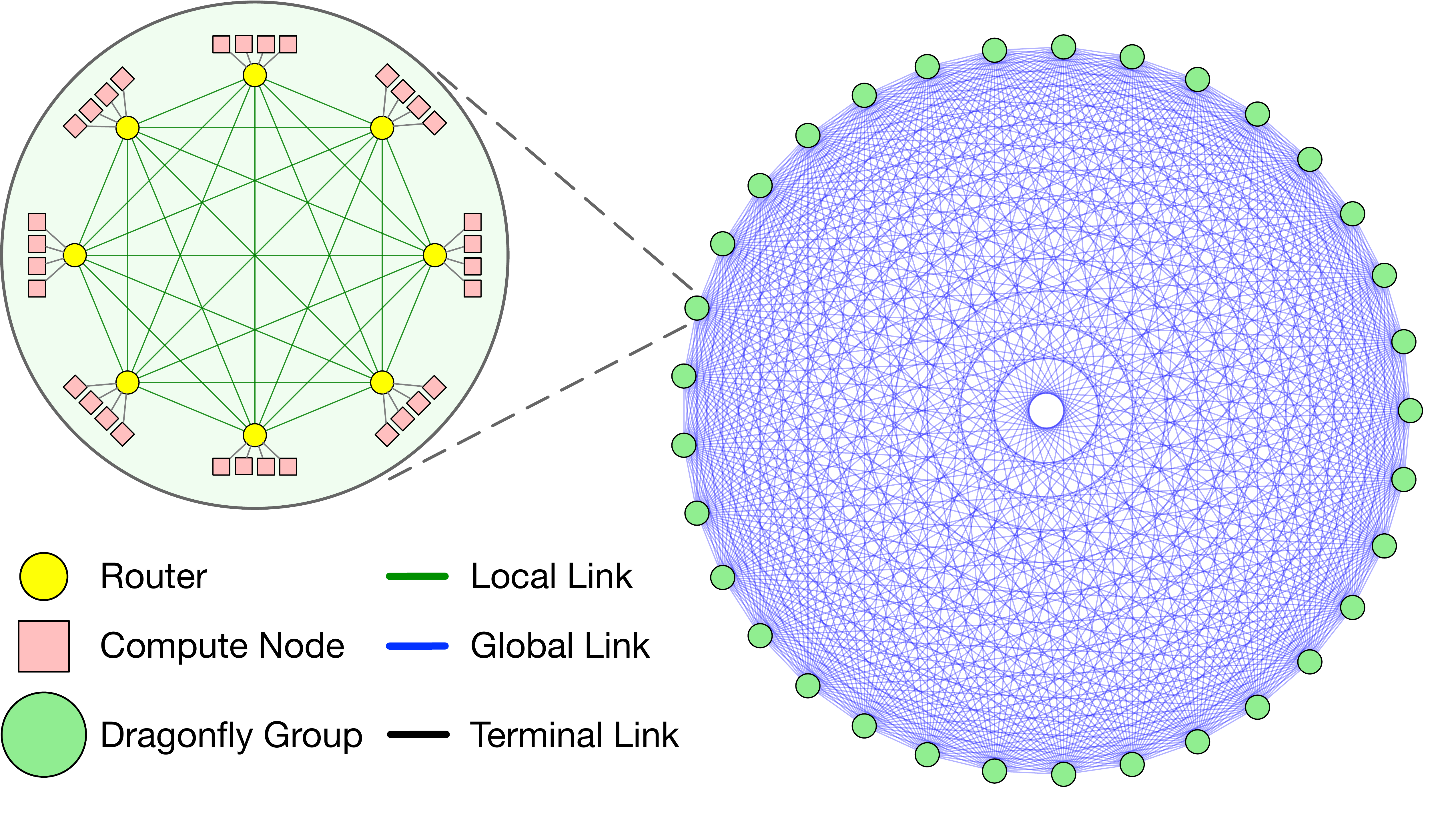}
    \caption{Topology of a 1,056-node 1D Dragonfly network.}
    \vspace{-2mm}
    \label{fig:dragonfly}
\end{figure}

\textbf{Dragonfly Network Topology.} The Dragonfly topology is widely adopted in exascale HPC systems, including those in the TOP500 list \cite{top500}. It offers high bandwidth and low diameter while balancing scalability and cost. As shown in Figure \ref{fig:dragonfly}, Dragonfly \cite{dragonfly} is organized into groups of routers connected via local links, with compute nodes attached through terminal ports. Groups are interconnected in an all-to-all pattern using global links, enabling low-hop communication across the system.
We focus on the 1D Dragonfly variant, where routers within each group are fully connected. This design underlies Slingshot networks used in production systems such as Frontier and Aurora \cite{frontier, aurora}, making it a relevant target for our study.

\textbf{Hyrbid Network Simulation.} Parallel discrete event simulation (PDES) is widely used for high fidelity modeling of large scale network systems. CODES, built on the ROSS engine, enables flit level simulation and has been applied to various HPC interconnect studies \cite{mubarak2017enabling, yang2016watch, yang2016study, wolfe2017preliminary}.

\begin{figure}[ht]
    \centering
    \includegraphics[width=0.8\columnwidth]{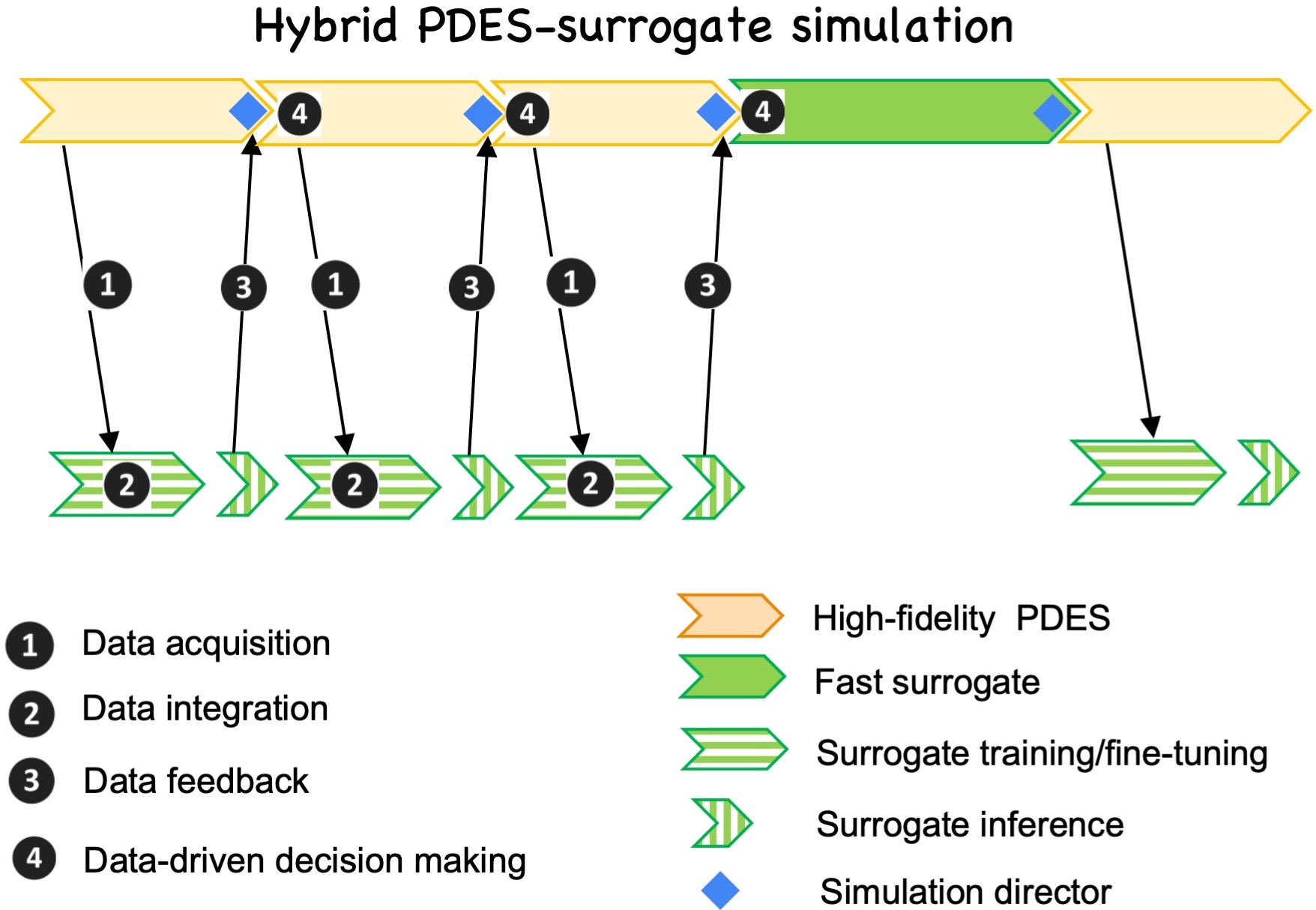}
    \caption{
    Hybrid PDES-surrogate simulation \cite{dddas}. }
    \label{fig:dddas-plot}
    \vspace{-2mm}
\end{figure}

Due to the high computational cost of PDES, hybrid approaches combine it with machine learning surrogate models to improve scalability \cite{elkin-tomacs}. Figure~\ref{fig:dddas-plot}, shows this process with four stages: (1) simulation data is collected, (2) used to train or fine tune the surrogate model, (3) the surrogate is validated through real time PDES output, and (4) a control loop decides whether to continue with PDES or switch to the faster surrogate. This adaptive feedback mechanism enables accurate yet efficient simulation.

\section{Related Work}
\textbf{Application runtime forecasting. }
Time series prediction has been widely studied in statistics and machine learning, with techniques such as ARIMA (Autoregressive Integrated Moving Average) \cite{arima}, Multi-Layer Perceptrons \cite{mlp1}, Convolutional Neural Networks \cite{cnn}, RNNs (Recurrent Neural Networks) \cite{rnn}, and LSTMs (Long Short-Term Memory) \cite{lstm}. These methods primarily focus on temporal data, often neglecting the spatial dependencies inherent in systems like HPC networks. Moreover, graph neural networks (GNNs) have been used to estimate metrics such as delay, jitter, and packet loss, as well as to enhance modeling expressiveness and granularity \cite{suarez2022graph, ferriol2023routenet, wang2022xnet}. A recent effort \cite{dddas} applied the Diffusion Convolutional Recurrent Neural Network (DCRNN) to predict application iteration times on a 72-node Dragonfly system. Different from this, we propose a novel integration of graph neural networks and large language models (LLMs) for more accurate predictions.

\textbf{Large Language Models.}
Large Language Models (LLMs) have recently gained significant attention for their ability to handle a wide range of tasks, including the ones on time-series data. More specifically, LLMs have been explored for their potential to capture complex temporal patterns without requiring fine-tuning. 
Recent studies \cite{timesfm, timellm, chen2024llmtsintegratorintegratingllm, garza2024timegpt1} have demonstrated the effectiveness of LLMs in time series prediction by leveraging their inherent ability to model sequential data and their capacity to generalize across domains. In contrast, our approach integrates LLMs and GNNs to capture both temporal and spatial dependencies in HPC systems simulations.
\textbf{Graph Neural Networks (GNNs).}
GNNs have emerged as a powerful class of models for learning over graph-structured data. Differently from traditional deep learning models that work on data in the Euclidean domain, GNNs can handle irregular data that can be represented with a graph. Examples are social networks, molecular graphs, and communication networks. GNNs operate by aggregating information from a node's neighbors: this enables them to capture both local and global structural dependencies and makes them effective on relational data. One of the approaches is the GraphSAGE \cite{graphsage}, which generalizes GNNs to large-scale graphs by sampling and aggregating features from a node's local neighborhood. Another popular model is the Graph Attention Network (GAT) \cite{veličković2018graphattentionnetworks}, which introduces attention mechanisms to weigh the importance of neighboring nodes dynamically.
Recent advances in GNNs have led to the development of models that can handle graph structures that evolve over time. These models usually combine GNNs with sequential models, such as RNNs \cite{rnn} or Transformers \cite{attention}.
For example, the Temporal Graph Network (TGN) \cite{rossi2020temporalgraphnetworksdeep} extends GNNs to handle continuous-time dynamic graphs, making it suitable for applications like social network analysis and recommendation systems.

\textbf{Graph-based Methods for Application Runtime Forecasting.} 
Graph-based learning has gained traction in network performance modeling, enabling fine-grained representations of network structure and dynamics. Previous studies have used GNNs to estimate flow level metrics \cite{suarez2022graph, ferriol2023routenet, wang2022xnet}, often in the context of generic communication or datacenter networks.
Our focus is on Dragonfly, an interconnect network commonly used in HPC systems, that employs adaptive routing that dynamically adjusts paths based on network conditions. Because of this, network interference and application variability are significantly more complex and larger in Dragonfly networks, which motivates the need for specialized modeling in this study.
A step toward this goal was taken by applying DCRNN to runtime prediction \cite{dddas}, but the effectiveness of the model diminishes when scaled beyond small networks. Moreover, recent works such as PraNet \cite{pranet} model short-term traffic bursts in generic internet or metaverse simulations using GNNs, transformers, and queuing theory with limited simulation data.
In contrast, \ourmodel focuses on high-fidelity simulation traces from HPC interconnects and proposes the first integration of GNNs and LLMs for spatio-temporal modeling of application runtime. This combination enables the model to generalize across job placement strategies, workloads, and dynamic traffic patterns, offering practical accuracy and inference efficiency at scale.

SMART advances this direction by combining GNNs with LLMs to jointly model spatial structure and temporal dynamics from port-level simulation data. To our knowledge, this is the first surrogate model integrating GNNs and LLMs for large-scale HPC runtime forecasting.

\section{Our Methodology}
\subsection{Networking Data}
High-fidelity simulations generate two main datasets: (1) \textit{application data}, which captures workload characteristics, and (2) \textit{router data}, which records port-level features every \SI{250}{\micro\second}. These are used to construct a \emph{temporal graph} representation.

The temporal graph is represented as a sequence \( G_1, G_2, \dots, G_T \), where each graph \( G_i \) corresponds to iteration \( i \). The node set \( V \) and edge set \( E \) remain fixed, while node features change over time.
In the graphs, \emph{nodes} represent the router ports of the system, with some ports also representing the computing devices directly connected to them in the Dragonfly topology. \emph{Edges} are defined in two ways: (1) fully connecting nodes accounting for ports of the same router and (2) connecting ports connected by global or local links. This results in a complete graph structure that captures the connectivity of a dragonfly system.

\textbf{Node features.}
Each node is annotated with features that describe the network state at the corresponding port during each iteration. Since network snapshots are taken every \SI{250}{\micro\second}, we aggregate the snapshots over the iteration period by statistics like minimum, maximum, average or a quantile.

\emph{Active nodes} are Dragonfly computing units executing workloads whose iteration times are being predicted. For these nodes, aggregation bounds are defined by the start and end timestamps of the iteration. For non-active nodes, the aggregation interval $[lb, ub)$ is determined by: (1) if one of the terminal ports is active, the iteration timestamps of that node define the bounds, (2) if all terminal ports are active, the bounds are set by the earliest start and the latest end timestamps across the iterations, (3) if none of the terminal ports is active, the bounds are derived from the nearest active node's timestamps.

\subsection{Problem Definition}
Let the Dragonfly network consist of $n_r$ routers, $N_c$ computer nodes, and application has $N_p$ processes distributed across the computer nodes for execution. During execution, the $N_p$ processes collaborate to complete $T$ iterations. For each process $p \in \{1, 2, \dots, N_p\}$ at a specific iteration $t \in \{1, 2, \dots, T\}$, we define $y_{p,t}$ as the time taken for its application iteration and $x_{p,t}$ as the network characteristics of the router connected to the compute node hosting the rank $p$. To simplify the notation, we drop the subscript $p$ in the following descriptions. We formally define the problem below. 

\textbf{Problem Statement.}  Given a sequence of application iteration times $y$ and network characteristics $x$ for process rank $p$ within a look-back window, we define $\mathcal{B}$ as the input:
\begin{align*}
\mathcal{B} = \{ & y_{t-(L_y-1)}, y_{t-(L_y-2)}, \dots, y_{t}; \\
                 & x_{t-(L_x-1)}, x_{t-(L_x-2)}, \dots, x_{t} \}
\end{align*}
consisting of look-back windows of length $L_y$ for $y$ and $L_x$ for $x$. 
The goal is to predict the future application iteration times at $t+1$, denoted as $y_{t+1}$.

\subsection{Our Model: \ourmodel}

\begin{figure}[ht]
    \centering
    \includegraphics[width=.9\columnwidth]{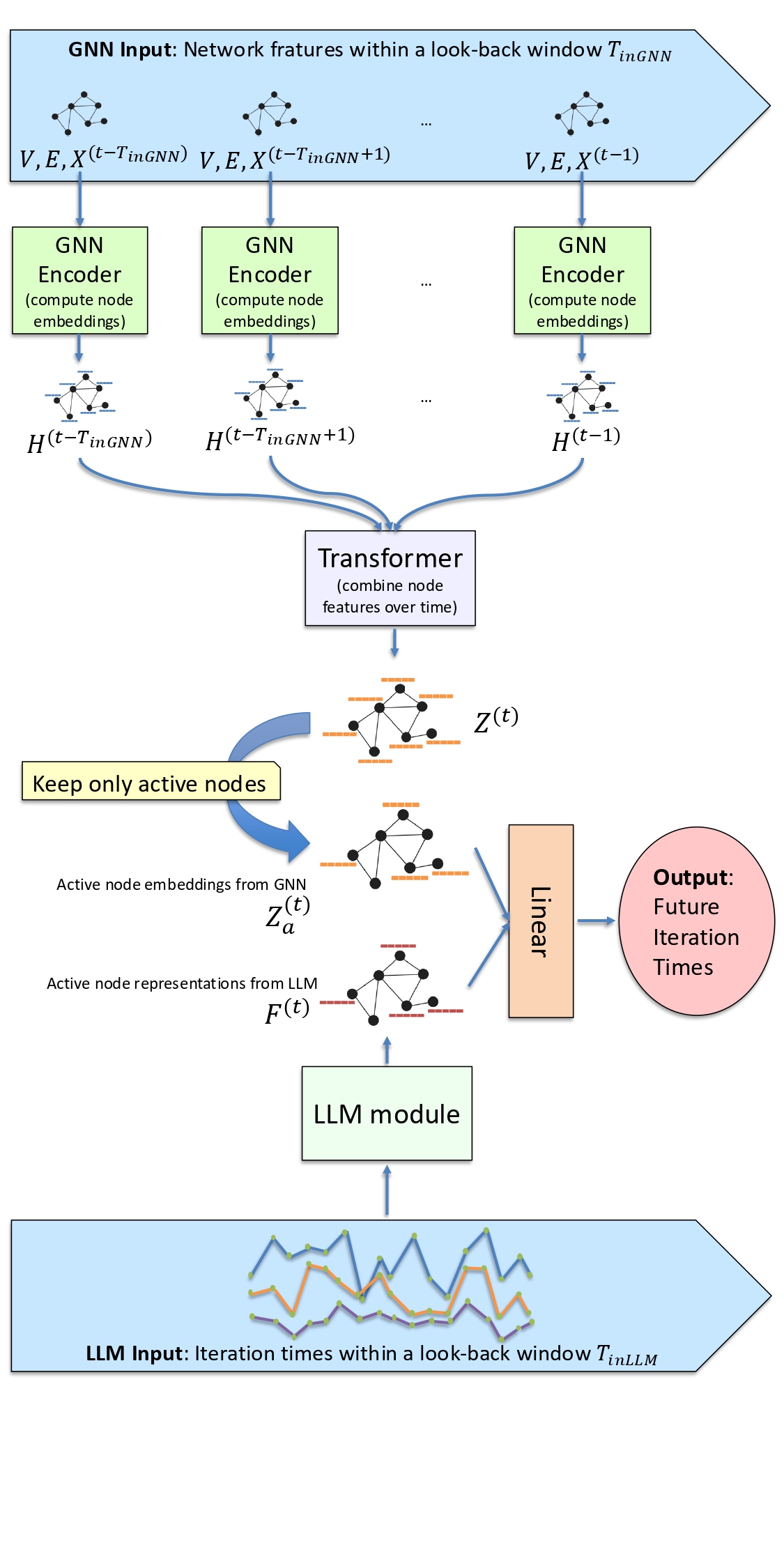}
    \vspace{-20mm}
    \caption{The architecture of our proposed model, \ourmodel. The GNN Encoder (top) generates the node embeddings while the Transformer helps to produce node representations over time. The lower part shows the LLM-based component. The outputs of each component is concatenated node-wise and fed into a Linear layer for the final prediction. }
    \label{fig:model-arch}
    \vspace{-4mm}
\end{figure}

Our model, \ourmodel, has three components, GNN, Transformer, and LLM, integrated for spatio-temporal modeling of workload contention and temporal variation in Dragonfly systems, which conventional graph or time series models cannot capture. Figure \ref{fig:model-arch} shows the architecture.
Table \ref{tab:notations} summarizes the notations.

    \begin{table}[t]
    \centering
    \caption{Notations used in the model description}
    \begin{adjustbox}{max width = \columnwidth}
    \begin{tabular}{ll}
        \toprule
        Symbol & Description \\
        \midrule
        $G_t$ & Graph representing the system at iteration $t$ \\
        $V$ & Set of all nodes (router ports) in the graph \\
        $V_a$ & Subset of $V$ corresponding to active nodes \\
        $E$ & Set of edges (connections between router ports) \\
        $X^{(t)}$ & Node feature matrix at iteration $t$: $(|V| \times d_f)$ \\
        $y_t$ & Application iteration times for iteration $t$ \\
        $T_{inGNN}$ & Length of the look-back window for GNN input \\
        $T_{inLLM}$ & Length of the look-back window for LLM input \\
        $H^{(t)}$ & Node embedding matrix from the GNN encoder: $(|V| \times d_h)$ \\
        $Z^{(t)}$ & Temporal node embeddings from the Transformer:$(|V| \times d_z)$ \\
        $E^{(t)}$ & LLM-generated embeddings for active nodes: $(|V_a| \times d_{llm})$ \\
        $F^{(t)}$ & Reduced dimensionality LLM embeddings  $(|V_a| \times d_z)$ \\
        $d_f$ & Number of features per node in $X^{(t)}$ \\
        $d_h$ & Dimensionality of GNN embeddings \\
        $d_z$ & Dimensionality of temporal embeddings from Transformer \\
        $d_{llm}$ & Dimensionality of LLM hidden state \\

        \bottomrule
    \end{tabular}
    \label{tab:notations}
    \end{adjustbox}
    
\end{table}

\subsubsection{GNN Encoder}
Our first goal is to capture the topological aspect of the dragonfly network with a graph neural network (GNN)-based encoder. The GNN encoder takes a sequence of temporal graphs \( \{G_{t-T_{inGNN}}, G_{t-T_{inGNN}+1}, \dots, G_{t-1}\} \) within a look-back window of length $T_{inGNN}$ where each graph \( G_t = (V, E, X^{(t)}) \) represents the network state at iteration \( t \). Here, \( V \) is the set of nodes (router ports), \( E \) is the set of edges (connections between ports), and \( X^{(t)} \in \mathbb{R}^{|V| \times d_f} \) is the node feature matrix at iteration \( t \), with \( d_f \) being the number of features per node. The GNN Encoder is a Graph Convolutional Network (GCN) \cite{gcn} which is applied to each graph \( G_t \) to capture individual spatial dependencies. The propagation rule for layer \( l+1 \) in GCN is given by:

\begin{equation*}
H^{(l+1, t)} = \sigma\left(\tilde{D}^{-1/2} \tilde{A} \tilde{D}^{-1/2} H^{(l, t)} W^{(l)}\right)
\end{equation*}
where \(l\) is the layer, \(t\) is the iteration number, \(\tilde{A} = A + I_N\) is the adjacency matrix \(A\) with self-loops, \(\tilde{D}\) is the diagonal degree matrix of \(\tilde{A}\), \(H^{(l)}\) represents the node feature matrix at layer \(l\), with \(H^{(0, t)}  = X^{(t)}\), \(W^{(l)}\) is the trainable weight matrix, \(\sigma(\cdot)\) denotes the RELU activation function.
The output of the GNN Encoder is a sequence of node embeddings that capture the spatial dependencies in the network at each input iteration: \( \{H^{(1)}, H^{(2)}, \dots, H^{(T_{inGNN})}\} \), where \( H^{(t)} \in \mathbb{R}^{|V| \times d_h} \) is the node embedding matrix at iteration \( t \), and \( d_h \) is the dimensionality of the embeddings.

\subsubsection{Temporal transformer to encode temporal dependencies}
Our next goal is to capture the temporal dependencies in a unified way. \ourmodel utilizes a transformer \cite{attention} module to process the temporal graph embeddings \( \{H^{(1)}, H^{(2)}, \dots, H^{(T_{inGNN})}\} \) generated by the GNN encoder. Unlike traditional recurrent or convolutional neural networks, transformers are 
designed for sequence handling via self-attention. Here we employ transformer to capture the temporal dependencies in the graph embeddings from each iteration. The input to the transformer encoder is a sequence of node embeddings over time, enriched with positional encoding. The transformer module processes this input and applies an attention mechanism.

\underline{Time-dependent Attention.} The core operation of the transformer module in \ourmodel is the attention mechanism, which computes the importance of different parts of the input sequence. The input embedding of size ($T_{inGNN}, d_m$), where $T_{inGNN}$ is the length of the input sequence and $d_m (=d_h\times |V|)$ is its dimensionality, is linearly transformed into three matrices: Query ($Q$), Key ($K$), and Value ($V$). 
These matrices are computed as:
\begin{equation*}
Q = H W^Q, \quad K = H W^K, \quad V = H W^V
\end{equation*}
where $H$ in our setting is as follows:
\begin{equation*}
H = [H^{(t-T_{inGNN})}, H^{(t-T_{inGNN}+1)}, \dots, H^{(t-1)}] 
\end{equation*}

and $H\in \mathbb{R}^{T_{inGNN} \times |V| \times d_h}$. Moreover, \( W^Q, W^K, W^V \) are trainable weight matrices. Each attention head calculates scaled dot-product attention as:
\begin{equation*}
\text{Attention}(Q, K, V) = \text{softmax}\left(\frac{QK^\top}{\sqrt{d_k}}\right)V
\end{equation*}
where $d_k$ is the dimensionality of $K$. Multi-head attention is concatenation of the outputs of multiple attention heads:

\begin{equation*}
\text{multihead}(Q, K, V) = \text{concat}(\text{head}_1, \dots, \text{head}_h)W^O
\end{equation*}
Here, $\text{head}_i = \text{Attention}(QW_i^Q, KW_i^K, VW_i^V)$, and $W_i^Q$, $W_i^K$, $W_i^V$, and $W_i^O$ are learnable weight matrices.  This mechanism allows the transformer to focus on different parts of the sequence simultaneously to capture diverse dependencies.
The output of the temporal transformer is a sequence of node embeddings that encode both spatial---by the GNN encoder---and temporal dependencies within the input features look-back window. 

\underline{Output of the Transformer Module.} The final output from the encoder is passed to the decoder, together with the last temporal node embeddings. These embeddings represent the most recent state of the nodes and provide critical information for predicting the next temporal embedding. In the decoder, the target sequence (last temporal node embeddings) is shifted to the right, ensuring that the model does not access the embedding it is trying to predict. The shifted target and the output of the encoder are then fed into the second and third layers, which correspond to successive encoder blocks, where K and V are derived from the encoder outputs, and Q is derived from the target.
The output is finally passed to a dense layer in order to produce the temporal node embeddings.
The final output from the transformer is  \( Z^{(t)} \in \mathbb{R}^{|V| \times d_z} \).

\subsubsection{LLM-powered Forecasting} We utilize a large language model (LLM)-based approach to further enhance our model, \ourmodel. Unlike traditional sequence models, LLMs in \ourmodel enable contextual prompting, incorporating task-specific and domain knowledge alongside time-series data. This helps capture long-range dependencies and improves runtime prediction across diverse workloads.
To the best of our knowledge this is the first work that uses an LLM in the surrogate model to forecast application runtime. 
In \ourmodel, this component uses the historical application iteration time data in as a sequence \( \{y_{t-T_{inLLM}}, y_{t-T_{inLLM}+1}, \dots, y_t\} \) within a pre-defined look-back window $T_{inLLM}$.

We closely follow the mechanism of Time-LLM \cite{timellm}. The steps of this module is as follows. The input sequence is first patched to normalize the input
time series. Next we embed them via a \textit{patch embedding} module into a dimension that is compatible with the LLM via a
simple linear layer. The patches are then transformed via a multi-head attention layer to obtain the embeddings which are in the same space as the pre-trained LLM. Finally, the LLMs generate embeddings \( E^{(t)} \in \mathbb{R}^{|V_{a}| \times d_{llm}} \) to represent the temporal patterns in the iteration times. Here, $V_a$ represents the subset of $V$ corresponding to the active nodes and $d_{llm}$ is the number of hidden channels in the underlying LLM model.

\underline{Prompting. } We use \textit{Prompt-as-Prefix (PaP)} technique for prompting \cite{timellm}. In this technique, the patch embeddings are prefixed with prompts that contain context to the input, including task instructions, domain knowledge, and descriptive statistics about the time series. For example, statistics such as minimum, maximum, median values, and trends can be used for this purpose. We use the following prompt template given below, where $workload\_name$ is the name of the application workload, $T_{inLLM}$ is the length of the historical input window to the LLM component.

\begin{tcolorbox}[colback=yellow!10!white, colframe=red!50!black, title=Prompt for Forecasting]
The dataset contains application iteration times for the
{\textcolor{blue}{[$workload\_name$]}} workload.

Task description: Forecast the next step given the
previous {\textcolor{blue}{[$T_{inLLM}$]}} steps information.

Input statistics: min value [...], max value [...],
median value [...]
\end{tcolorbox}
\vspace{2mm}

The input---which is composed of the prefix prompt and embedded patches---fed into the LLM. The obtained LLM output \( E^{(t)} \in \mathbb{R}^{|V_{a}| \times d_{llm}} \) is flattened and linearly projected into an hidden dimension that matches the temporal node embeddings size from the transformer module, resulting in a final output \( F^{(t)} \in \mathbb{R}^{|V_{a}| \times d_{z}} \).

\subsubsection{Final Integration}

In the final step, we integrate GNN embeddings along with the LLM output effectively. The temporal embeddings \( Z^{(t)} \in \mathbb{R}^{|V| \times d_z} \) from the Transformer are filtered by applying a node mask to ensure that only active nodes---such as those with target values---are considered in the merging process. The result of this operation is \( Z_{a}^{(t)} \in \mathbb{R}^{|V_a| \times d_z} \). For each active node, the reduced-dimensional hidden state from the LLM is concatenated with the corresponding node embedding from the GNN. This creates a combined feature vector that encodes both spatial and temporal information powered by the LLM. The combined feature vector is passed through a fully connected layer to produce the final predictions ($\hat{y}_{t+1}$) for each node.Rather than manually tuning the balance between spatial and temporal modeling, SMART learns this integration end-to-end. The final prediction layer combines GNN-derived spatial features with LLM-derived temporal context, allowing the model to automatically determine their relative importance. 

\subsubsection{Online Tuning Strategy}

We develop an online tuning strategy to address the evolving nature of network traffic and workload behavior in Dragonfly systems. This approach is designed for hybrid simulation settings, where ground-truth data from PDES is intermittently available. The model is fine-tuned during inference only when such feedback is present, rather than assumed to be continuous. With a feedback loop, \ourmodel adjusts its parameters to align with the latest network dynamics, thus maintaining prediction accuracy over time. This strategy is particularly beneficial in environments where workloads are heterogeneous and exhibit temporal variability.
Our approach has two phases. The first one is the offline learning over a training dataset consisting of 30\% of the available data. The second phase is the inference phase. Here we consider the ground truth data to be available for every $F_t$ iterations. This implies that if the iteration index $t$ is multiple of $F_t$, we use the data from $t - F_t$ to $t$ to update the model weights via the usual back-propagation. In the following, $F_t = \infty$ means that the online tuning strategy is turned off, i.e. the inference phase does not incorporate any update of the model weights. 
\section{Experiments}

\begin{table}[ht]
\centering
\caption{Model hyperparameters}
\label{tab:hyperparams}
\begin{adjustbox}{max width=\columnwidth}
\begin{tabular}{cc}
\toprule
\textbf{Parameter}  &  \textbf{Value} \cr
\midrule
Patch length in LLM & 2 \cr
Stride in LLM & 1 \cr
LLM hidden channels $D_{llm}$ & 768 \cr
LLM num hidden layers & 32 \cr
LLM model & GPT-2 \cr
GCN layers & 2 \cr
GCN embedding size $d_h$ & 128 \cr
Transformer model dimension $d_z$ & 128 \cr
Number of attention heads in Transformer & 8 \cr
Transformer encoder layers & 2 \cr
Transformer decoder layers & 2 \cr
\bottomrule
\end{tabular}
\end{adjustbox}
\end{table}

\begin{table*}[ht]
\centering
\caption{MAPE comparison of \ourmodel versus other baselines for different historical input windows $T_{inGNN}$, $T_{inLLM}$ and retrain frequencies $F_T$ over the simulation in $D_1$. Bold indicates the best result, underlined indicates the second best (column-wise). \ourmodel consistently outperforms the baselines.} 
\vspace{-1mm}
\label{tab:res1}
\begin{adjustbox}{max width=.9\textwidth}
\vspace{-2mm}
\begin{tabular}{cccccccccccccccc}
\toprule

\multicolumn{3}{c}{\textbf{\ourmodel}} & \multicolumn{12}{c}{$T_{inGNN}$} \\
\hline

\multicolumn{3}{c}{} & \multicolumn{4}{c}{2} & \multicolumn{4}{c}{4} & \multicolumn{4}{c}{8} \\
\hline

\multicolumn{3}{c}{} & \multicolumn{2}{c}{Cont} & \multicolumn{2}{c}{Rand} & \multicolumn{2}{c}{Cont} & \multicolumn{2}{c}{Rand} & \multicolumn{2}{c}{Cont} & \multicolumn{2}{c}{Rand} \\
\hline

\multicolumn{2}{c}{} 
& \makecell[tc]{\rotatebox{90}{\makecell{Tuning \\ frequency $F_t$ } }} 
& \makecell[tc]{\rotatebox{90}{MILC}} 
& \makecell[tc]{\rotatebox{90}{LAMMPS }} 
& \makecell[tc]{\rotatebox{90}{MILC}} 
& \makecell[tc]{\rotatebox{90}{LAMMPS}} 
& \makecell[tc]{\rotatebox{90}{MILC}} 
& \makecell[tc]{\rotatebox{90}{LAMMPS}} 
& \makecell[tc]{\rotatebox{90}{MILC}} 
& \makecell[tc]{\rotatebox{90}{LAMMPS}} 
& \makecell[tc]{\rotatebox{90}{MILC}} 
& \makecell[tc]{\rotatebox{90}{LAMMPS}} 
& \makecell[tc]{\rotatebox{90}{MILC}} 
& \makecell[tc]{\rotatebox{90}{LAMMPS}} \\

\hline

\multirow{6}{*}{\rotatebox{90}{$T_{inLLM}$}} & \multirow{2}{*}{2} & $\infty$ & 3.98 & 2.64 & 3.9 & 2.54 & 4.30 & 2.64 & 4.17 & 2.59 & 4.64 & 2.88 & 4.30 & 2.67 \\

 &  & 8 & 3.70 & 2.51 & 3.81 & 2.47 & 3.91 & 2.62 & 4.04 & 2.53 & 4.36 & 2.80 & 4.12 & 2.60 \\

 & \multirow{2}{*}{4} & $\infty$ & 3.87 & 1.90 & 3.72 & 1.81 & 4.22 & \underline{1.94} & 3.91 & \underline{1.85} & 4.28 & 2.04 & 3.96 & \underline{1.90} \\

 &  & 8 & 3.75 & \underline{1.86} & \underline{3.38} & \underline{1.78} & 4.05 & \textbf{1.92} & \underline{3.52} & \textbf{1.82} & 4.18 &  \textbf{1.97} & \underline{3.59} & \textbf{1.88} \\

 & \multirow{2}{*}{8} & $\infty$ & \underline{3.50} & 1.89 & 3.40 & 1.87 & \underline{4.04} & 2.06 & 3.67 & 1.89 & \underline{3.90} & 2.05 & 3.71 & 1.95 \\

 &  & 8 & \textbf{3.19} & \textbf{1.78} & \textbf{3.12} & \textbf{1.84} & \textbf{3.76} & 1.96 & \textbf{3.34} & 1.86 & \textbf{3.74} & \underline{2.03} & \textbf{3.44} & 1.91 \\
\midrule
\multicolumn{3}{c}{\textbf{Baselines}} 
 &  &  &  &  &  &  &  &  & &  &  & \\
 \hline
 
\multicolumn{2}{c}{\multirow{2}{*}{LSTM} }
 & $\infty$ & 6.15 & 7.29 & 6.11 & 7.32 & 5.90 & 7.20 & 5.89 & 7.16 & 5.90 & 7.20 & 5.88 & 6.54 \\  
\multicolumn{2}{c}{} & 8 & 6.09 & 7.25 & 6.05 & 7.28 & 5.82 & 7.15 & 5.81 & 7.11 & 5.79 & 6.48 & 5.83 & 6.53 \\ \hline

\multicolumn{2}{c}{\multirow{2}{*}{DCRNN}}
 & $\infty$ & 6.27 & 7.14 & 6.04 & 7.13 & 6.07 & 7.36 & 6.07 & 7.35 & 6.11 & 6.53 & 5.99 & 6.86 \\  
\multicolumn{2}{c}{} & 8 & 6.35 & 7.10 & 5.98 & 7.10 & 5.97 & 7.34 & 5.97 & 7.33 & 6.00 & 6.49 & 5.88 & 6.82 \\ \hline

\multicolumn{3}{c}{LAST} 
 & 7.86 & 8.14 & 7.83 & 8.21 & 7.86 & 8.14 & 7.83 & 8.21 & 7.86 & 8.14 & 7.83 & 8.21 \\

\hline

\multicolumn{3}{c}{MEAN} 
 & 7.80 & 8.24 & 9.31 & 7.72 & 10.37 & 12.85 & 10.96 & 13.14 & 12.86 & 13.11 & 13.15 & 13.53 \\
\bottomrule

\end{tabular}

\end{adjustbox}
\vspace{-2mm}
\end{table*}

We demonstrate the effectiveness of \ourmodel in predicting application runtime on a 1,056 node Dragonfly system (Figure~\ref{fig:dragonfly}), where multiple applications execute concurrently and contend for shared network resources. Our evaluation examines prediction quality across a range of workloads and placement policies, and measures the inference time overhead to confirm that \ourmodel is suitable for integration into hybrid simulation workflows. We further conduct ablation studies to quantify the contributions of the GNN and Time LLM components, analyze model sensitivity through hyperparameter variation, and compare \ourmodel against several state of the art temporal forecasting models. 
The code and datasets used in this study are available as noted after the abstract.

\subsection{Experimental Setup}

\subsubsection{Data Generation.}
We analyze two datasets from CODES simulations that capture the execution of multiple HPC applications running concurrently on a 1,056-node Dragonfly system \cite{codes-git}. 
We select several representative HPC applications spanning computational physics, molecular dynamics, and structured grid-based computations.  

\textbf{Dataset $D_1$: }The first dataset ($D_1$) includes two application workloads: MILC \cite{milc} and LAMMPS \cite{lammps}. In this dataset, 512 nodes run  MILC, another 512 nodes run LAMMPS, and the remaining 36 nodes generate Uniform Random (UR) background traffic, where each node sends successive messages to random destinations. MILC is an HPC application used for quantum chromodynamics (QCD) simulations, while LAMMPS is a molecular dynamics code focused on materials modeling.  

\textbf{Dataset $D_2$: }The second dataset ($D_2$) extends $D_1$ by adding a third workload, NN, which represents a 27-point 3D stencil computation used for modeling iterative updates over a structured grid. Stencil computations are a fundamental pattern in HPC workloads, commonly used in scientific simulations, computational fluid dynamics, and iterative solvers. In this setup, 384 nodes run MILC, 512 nodes run LAMMPS, and 160 nodes run NN.  

\textbf{Job placement policies.} Since application performance is significantly affected by different job placement policies\cite{yang2016watch}, we consider two widely used strategies for each dataset: \emph{contiguous} and \emph{random} job placement. In the random placement strategy, workload-assigned compute nodes are distributed across the system, whereas in the contiguous placement strategy, they are allocated sequentially. Thus, we evaluate our model using two datasets, each with two different job placement policies. 

Consequently, our experiments encompass four different application and job placement combinations: \textbf{D1-Rand} (Dataset D1 under random placement), \textbf{D1-Cont} (Dataset D1 under contiguous placement), \textbf{D2-Rand} (Dataset D2 under random placement), and \textbf{D2-Cont} (Dataset D2 under contiguous placement).

The $D_1$ simulations were conducted on a Bebop Broadwell node \cite{bebop}, equipped with 36 Intel Xeon E5-2695v4 CPUs and 128GB of memory, with execution times of 66.17 hours for random placement and 38.68 hours for contiguous placement. 
The $D_2$ simulations were run on a compute\_icelake\_r650 machine on Chameleon\cite{keahey2020lessons}, featuring two Intel(R) Xeon(R) Platinum 8380 CPUs @ 2.30GHz and 256 GiB of memory, with execution times of 62.46 hours for random placement and 36.37 hours for contiguous placement.

\subsubsection{Baselines.}
We use the following baselines:
\begin{itemize}
    \item \textbf{Last} is a heuristic that forecasts the next iteration time as the last historical available one, i.e. $ \hat{y}_t=\hat{y}_{t+1} = \hat{y}_{t+2} = ... = \hat{y}_{t+T_{out}} = y_{t-1}$.
    
    \item \textbf{Mean} forecasts based on the past W average iteration times, i.e. $\hat{y}_t=\hat{y}_{t+1} = \hat{y}_{t+2} = ... = \hat{y}_{t+T_{out}} = \frac{\sum_{i=t-W}^{t-1} y_i}{W}$. In the experiments, we use $W = T_{inGNN}$.

    \item \textbf{LSTM} (Long-Short Term Memory \cite{lstm}) predicts the next iteration time based on the historical iteration time values within a look-back window varied according to $T_{inGNN}$. 

    \item \textbf{DCRNN-based Baseline.} This recent method \cite{dddas} uses a popular GNN-based architecture named as Diffusion Convolutional Recurrent Neural Network (DCRNN). To have a fair comparison, the length of the look-back window is varied according to $T_{inGNN}$.
\end{itemize}

\subsubsection{Performance Mesaures.} \
To evaluate the performance of the models, we consider two key measures: (1) forecasting quality and (2) inference time. We use the standard Mean Absolute Percentage Error (MAPE) as the metric to assess quality, which is computed as: $ \text{MAPE} = \frac{1}{N} \sum_{i=1}^{N} \left| \frac{y_i - \hat{y}_i}{y_i} \right| * 100
$ where $y_i$ represents the actual iteration time, and $\hat{y}_i$ represents the predicted iteration time.

\subsubsection{Other Settings. }For each application, dataset and job placement strategy, we train and test the models tailored to predict the specific application iteration times. 
Table \ref{tab:hyperparams} shows all the hyperparameter values that are used in our experiments. 
We run the experiments for different combinations of $T_{inLLM} = 2, 4, 8$ and $T_{inGNN} = 2, 4, 8$ and tuning frequencies $F_T = \infty, 8$, where $F_T = \infty$ means that the model is never tuned during the testing, while $F_T = 8$ means that every 8 test iterations, the model weights are updated to allow the model to capture recent changes in the network patterns.
Training each SMART model takes approximately 3–4 hours using two NVIDIA A100-PCIE GPUs (40 GiB memory each) with CUDA 12.4. Training is a one-time cost, while inference is lightweight and efficient, as detailed in Table 2 of the main paper.
All reported results are based on a single run per experimental setting.

\begin{table*}[ht]
\centering
\caption{MAPE comparison for different historical input windows $T_{inGNN}$, $T_{inLLM}$ and retrain frequencies $F_T$ over the simulation in $D_2$. Bold indicates the best result, underlined indicates the second best (column-wise). \ourmodel consistently outperforms the baselines; the best result is often obtained with longer LLM input window ($T_{inLLM} = 8$).}
\label{tab:res2}
\vspace{-2mm}
\begin{adjustbox}{max width=\textwidth}

\begin{tabular}{ccccccccccccccccccccc}
\toprule

\multicolumn{3}{c}{\textbf{\ourmodel}} & \multicolumn{18}{c}{$T_{inGNN}$} \\
\hline

\multicolumn{3}{c}{} & \multicolumn{6}{c}{2} & \multicolumn{6}{c}{4} & \multicolumn{6}{c}{8} \\
\hline

\multicolumn{3}{c}{} & \multicolumn{3}{c}{Cont} & \multicolumn{3}{c}{Rand} & \multicolumn{3}{c}{Cont} & \multicolumn{3}{c}{Rand} & \multicolumn{3}{c}{Cont} & \multicolumn{3}{c}{Rand} \\
\hline

\multicolumn{2}{c}{} 
& \makecell[tc]{\rotatebox{90}{\makecell{Tuning \\ frequency $F_t$ } }} 
& \makecell[tc]{\rotatebox{90}{MILC}} 
& \makecell[tc]{\rotatebox{90}{LAMMPS }} 
& \makecell[tc]{\rotatebox{90}{NN}} 
& \makecell[tc]{\rotatebox{90}{MILC}} 
& \makecell[tc]{\rotatebox{90}{LAMMPS}} 
& \makecell[tc]{\rotatebox{90}{NN}} 
& \makecell[tc]{\rotatebox{90}{MILC}} 
& \makecell[tc]{\rotatebox{90}{LAMMPS}} 
& \makecell[tc]{\rotatebox{90}{NN}} 
& \makecell[tc]{\rotatebox{90}{MILC}} 
& \makecell[tc]{\rotatebox{90}{LAMMPS}} 
& \makecell[tc]{\rotatebox{90}{NN}} 
& \makecell[tc]{\rotatebox{90}{MILC}} 
& \makecell[tc]{\rotatebox{90}{LAMMPS}} 
& \makecell[tc]{\rotatebox{90}{NN}} 
& \makecell[tc]{\rotatebox{90}{MILC}} 
& \makecell[tc]{\rotatebox{90}{LAMMPS}} 
& \makecell[tc]{\rotatebox{90}{NN}} \\
\hline

\multirow{6}{*}{\rotatebox{90}{$T_{inLLM}$}} & \multirow{2}{*}{2} & $\infty$ & 4.29 & 2.81 & 3.30 & 4.04 & 2.63 & 3.05 & 4.70 & 3.04 & 3.49 & 4.24 & 2.71 & 3.29 & 4.92 & 3.14 & 3.69 & 4.37 & 2.82 & 3.39 \\
 &  & 8 & 4.12 & 2.72 & 3.21 & 3.98 & 2.58 & 2.97 & 4.39 & 2.95 & 3.30 & 4.18 & 2.63 & 3.24 & 4.75 & 3.02 & 3.71 & 4.28 & 2.71 & 3.33 \\
& \multirow{2}{*}{4} & $\infty$ & 4.03 & 1.99 & 3.21 & 3.92 & 1.92 & 2.88 & 4.64 & \textbf{2.13} & \textbf{3.19} & 4.08 & {1.97} & 2.97 & 4.63 & \underline{2.20} & 3.40 & 4.12 & {2.05} & 3.06 \\
 &  & 8 & 4.11 & \underline{1.98} & \underline{2.89} & 3.83 & {1.90} & \underline{2.85} & 4.59 & {2.23} & {3.26} & 3.96 & {1.96} & \underline{2.91} & 4.60 & {2.28} & {3.34} & 4.08 & {1.99} & \underline{2.97} \\
& \multirow{2}{*}{8} & $\infty$ & \underline{3.91} & 1.93 & 3.07 & \underline{3.51} & \underline{1.89} & \underline{2.85} & \underline{4.17} & 2.26 & \underline{3.22} & \underline{3.79} & \underline{1.91} & 2.93 & \underline{4.19} & \underline{2.20} & \underline{3.23} & \underline{3.87} & \underline{1.97} & 2.99 \\
 &  & 8 & \textbf{3.51} & \textbf{1.88} & \textbf{2.87} & \textbf{3.39} & \textbf{1.86} & \textbf{2.79} & \textbf{3.84} & \underline{2.20} & {3.30} & \textbf{3.63} & \textbf{1.88} & \textbf{2.82} & \textbf{4.17} & \textbf{2.14} & \textbf{3.22} & \textbf{3.74} & \textbf{1.93} & \textbf{2.90} \\

\hline
\multicolumn{3}{c}{\textbf{Baselines}} 
 &  &  &  &  &  &  &  &  & &  &  & \\
 \hline
\multicolumn{2}{c}{\multirow{2}{*}{LSTM}} & $\infty$ & 6.23 & 8.14 & 6.96 & 6.23 & 7.45 & 6.92 & 6.97 & 8.28 & 7.66 & 5.72 & 7.11 & 6.81 & 6.32 & 7.95 & 7.81 & 5.59 & 6.46 & 7.21 \\
\multicolumn{2}{c}{} & 8 & 6.28 & 8.02 & 7.38 & 6.17 & 7.41 & 6.82 & 6.85 & 8.80 & 8.12 & 5.65 & 7.06 & 6.76 & 6.37 & 8.02 & 7.55 & 5.51 & 6.40 & 7.15 \\
\hline
\multicolumn{2}{c}{\multirow{2}{*}{DCRNN}} & $\infty$ & 6.54 & 7.54 & 7.22 & 6.13 & 7.25 & 6.83 & 6.78 & 8.47 & 8.13 & 5.85 & 7.21 & 6.87 & 6.65 & 8.43 & 7.87 & 5.85 & 6.42 & 7.17 \\
\multicolumn{2}{c}{} & 8 & 6.79 & 7.32 & 7.49 & 6.07 & 7.21 & 6.79 & 7.16 & 8.06 & 7.65 & 5.75 & 7.20 & 6.80 & 6.57 & 8.31 & 7.50 & 5.74 & 6.38 & 7.10 \\
\hline

\multicolumn{3}{c}{LAST}
& 8.73 & 9.11 & 8.97 & 8.84 & 9.03 & 8.68
& 8.73 & 9.11 & 8.97 & 8.84 & 9.03 & 8.68
& 8.73 & 9.11 & 8.97 & 8.84 & 9.03 & 8.68 \\

\hline
\multicolumn{3}{c}{MEAN}
& 8.20 & 9.00 & 8.63 & 7.75 & 8.18 & 7.71
& 8.89 & 9.63 & 8.68 & 11.32 & 13.17 & 9.83
& 12.51 & 15.24 & 11.26 & 13.46 & 13.70 & 9.76 \\
\bottomrule

\end{tabular}
\end{adjustbox}
\end{table*}

\begin{table*}[ht]
\centering
\caption{Statistics of application iteration times and prediction errors. The table lists the maximum and minimum iteration times (in nanoseconds), the standard deviation of iteration times, as well as the MAPE and RMSE of the \ourmodel predictions.
Despite the significantly higher iteration time variability under random placement, \ourmodel maintains consistent accuracy across both placement strategies for all applications.
}
\label{tab:res4}
\begin{tabular}{clrrrrr}
\toprule
&  & \multicolumn{1}{r}{Max iter. time (ns)} & \multicolumn{1}{r}{Min iter. time (ns)} & \multicolumn{1}{r}{Std.} & \multicolumn{1}{r}{\ourmodel MAPE} & \multicolumn{1}{r}{\ourmodel RMSE} \\ 
\midrule
\multicolumn{1}{l|}{\multirow{3}{*}{\rotatebox{90}{Cont}}} & MILC   & 1582600.75 & 1549385.13  & 3089.45   & 3.51 & 0.00247 \\
\multicolumn{1}{l|}{}                            & LAMMPS & 1539606.96 & 1036303.24  & 130891.07 & 1.88 & 0.00122 \\
\multicolumn{1}{l|}{}                            & NN     & 322073.02  & 161752.20   & 19313.22  & 2.87 & 0.00272 \\ 
\hline
\multicolumn{1}{l|}{\multirow{3}{*}{\rotatebox{90}{Rand}}}  & MILC   & 2173786.95 & 1543215.38  & 100769.42 & 3.39 & 0.00259 \\
\multicolumn{1}{l|}{}                            & LAMMPS & 6561415.11 & 1252410.89  & 807331.12 & 1.86 & 0.00109 \\
\multicolumn{1}{l|}{}                            & NN     & 740699.14  & 163661.67   & 59688.04  & 2.79 & 0.00244 \\ 
\bottomrule
\end{tabular}
\end{table*}

\subsection{Results on Forecasting Quality}
\label{sec:exp_quality}

This section evaluates the forecasting quality of \ourmodel alongside baseline models for predicting iteration time. Table~\ref{tab:res1} and Table~\ref{tab:res2} present the MAPE values for experiments conducted on datasets \(D_1\) and \(D_2\), respectively. The columns ``MILC", ``LAMMPS" and ``NN" refer to models trained considering the nodes corresponding to the specific application as active. The columns ``Cont" and "Rand" indicate whether the dataset follows a contiguous or random placement strategy. The cells between ``$T_{inLLM}$" and ``$T_{inGNN}$" refer to the results for the proposed \ourmodel model, the other cells refer to the results for the baselines. 

The results show that the best performance is obtained for $T_{inLLM} = 8, T_{inGNN} = 2$ in all the cases. This suggests the LLM effectively captures long-term temporal patterns while the GNN focuses on recent spatial dynamics. 
In addition, the online tuning strategy presented for the DCRNN model in \cite{dddas} appears to be beneficial. Incorporating an online tuning strategy ($F_t=8$) consistently reduces forecasting errors compared to a static model ($F_t=\infty$). By updating the model weights every 8 iterations, \ourmodel adapts to recent changes in network behavior, resulting in lower MAPE values in both data sets and placement types.

Table \ref{tab:res4} summarizes key statistics on application iteration times and prediction errors for MILC, LAMMPS, and NN in the $D_2$ dataset under both contiguous and random job placement strategies. The table reports the maximum and minimum iteration times, the standard deviation, as well as the Mean Absolute Percentage Error (MAPE) and Root Mean Square Error (RMSE) of the \ourmodel predictions. 
LAMMPS exhibits a significantly larger variance in iteration time compared to MILC and NN. However, \ourmodel effectively adapts to these fluctuations. These detailed statistics further validate \ourmodel's robustness in handling execution time variability caused by network interference.
Additionally, the experimental results show minimal accuracy differences between contiguous and random placements: MILC exhibits a MAPE difference of 0.12\%, and LAMMPS shows an even narrower gap of 0.02\%. However, random placement introduces greater iteration time variability due to network interference compared with contiguous placement. These findings demonstrate the robustness of \ourmodel in predicting application runtime across placement strategies.

\textbf{Discussion.} 
Achieving $100\%$ accuracy in practice is unrealistic, particularly in highly dynamic networking environments of large-scale systems. In a related study, researchers demonstrated that a hybrid PDES using a MEAN-based surrogate model could achieve comparable simulation performance \cite{elkin-tomacs}. 

Our results show that \ourmodel significantly outperforms MEAN across various scenarios. 
Prediction accuracy varies across different applications, with LAMMPS exhibiting higher precision than MILC. This discrepancy arises from their distinct sensitivities to network latency: LAMMPS is moderately sensitive, while MILC, with frequent global communication, is highly latency-sensitive. As a result, their
sensitivity to application communication prediction also differs,
affecting the accuracy of surrogate models. \textit{Overall, our model outperforms in prediction accuracy by effectively incorporating domain-specific knowledge of both application characteristics and network architecture.}

\begin{table}[ht]
\centering
\caption{
\textit{Average inference time} (in seconds). Benchmarked on an 80-core ARM system with 2×A100 GPUs. All models, including \ourmodel, achieve inference times significantly lower than the simulation per-iteration times (Table~\ref{table:sim_time}).
}
\label{table:inference_time}
\small
\begin{tabular}{cc}
\toprule
\textbf{Model} & \textbf{Avg. Inference Time} \\
\midrule
\ourmodel & 0.5150 \\
LLM-only & 0.4158 \\
GNN-only & 0.0633 \\
MEAN & 0.00001 \\
LAST & $<$ 0.00001 \\
LSTM & 0.0398 \\
DCRNN & 0.0459 \\
\bottomrule
\end{tabular}
\vspace{-2mm}
\end{table}

\begin{figure*}[t]
    \centering
    \begin{subfigure}{0.38\textwidth}
        \centering
        \includegraphics[width=\linewidth]{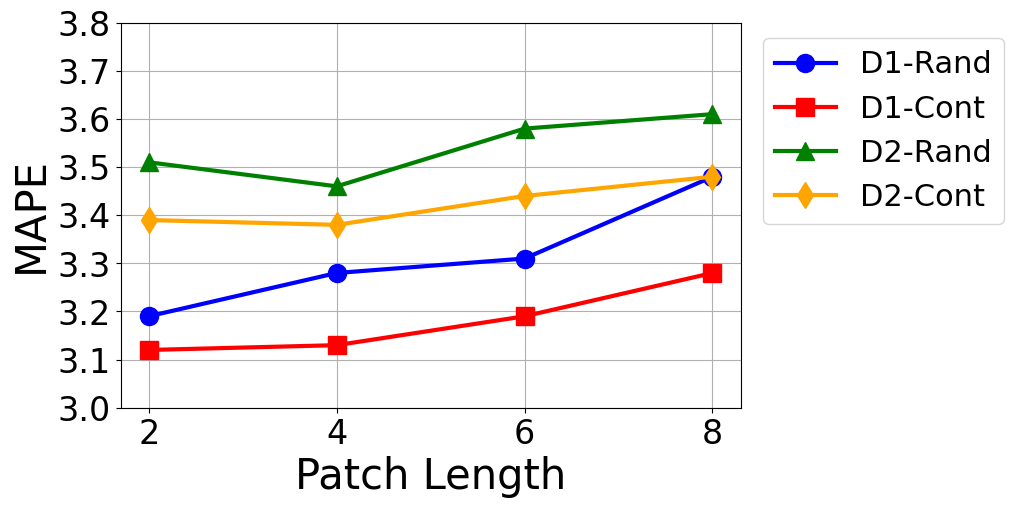}
        \caption{Effect of Patch Length}
        \label{fig:mape_patch_length}
    \end{subfigure}
    \begin{subfigure}{0.28\textwidth}
        \centering
        \includegraphics[width=\linewidth]{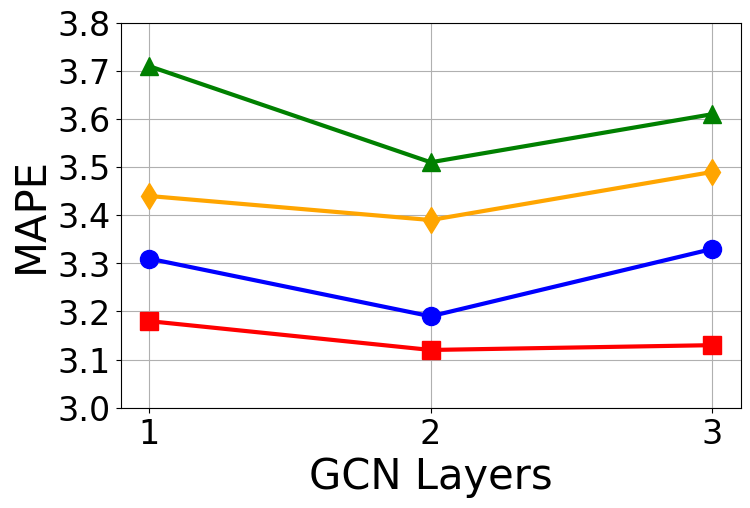}
        \caption{Effect of GCN Layers}
        \label{fig:mape_gcn_layers}
    \end{subfigure}
    \begin{subfigure}{0.28\textwidth}
        \centering
        \includegraphics[width=\linewidth]{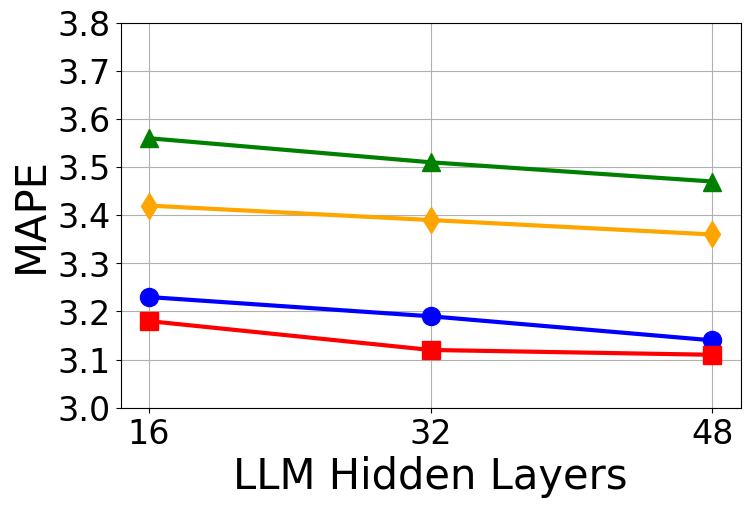}
        \caption{Effect of LLM Hidden Layers}
        \label{fig:mape_llm_layers}
    \end{subfigure}
    \caption{Effect of different hyper-parameters on MAPE for the different Datasets and node assignment combinations: (a) Path Length, (b) the number of GCN layers, and (c) the number of LLM hidden layer. The variations are not significant given that the MAPE varies between 3.1 and 3.8 the best baseline has a MAPE of 6.09. } 
    \label{fig:mape_comparison}
\end{figure*}

\begin{table}[t]
\centering
\caption{
\textit{Simulation time} per iteration (seconds) for each application in $D_1$ and $D_2$ under different job placements. All surrogate models (Table~\ref{table:inference_time}) are orders of magnitude faster, enabling integration with PDES.}
\label{table:sim_time}
\small
\begin{tabular}{clrr}
\toprule
Dataset  & Application & \multicolumn{1}{l}{Contiguous} & \multicolumn{1}{l}{Random} \\
\midrule
\multirow{2}{*}{$D_1$} & MILC   & 79.36  & 132.34   \\
                       & LAMMPS & 56.47  & 243.48   \\
\hline
\multirow{3}{*}{$D_2$} & MILC   & 65.46  & 112.42   \\
                       & LAMMPS & 47.06  & 216.58   \\
                       & NN     & 8.09   & 14.98    \\
\bottomrule
\end{tabular}
\vspace{-4mm}
\end{table}

\begin{table}[ht]
\centering
\caption{MAPE results for GNN-only on $D_1$ for different $T_{inGNN}$ and retrain frequencies $F_T$.}
\label{tab:gnn_only_d1}
\begin{adjustbox}{max width=\linewidth}
\begin{tabular}{c c | c c | c c}
\toprule
\multicolumn{2}{c|}{} &
\multicolumn{2}{c|}{Cont} &
\multicolumn{2}{c}{Rand} \\
\textbf{$T_{inGNN}$} & \textbf{$F_T$} &
MILC & LAMMPS &
MILC & LAMMPS \\
\midrule

\multirow{2}{*}{2}
 & $\infty$ & 3.68 & 2.14 & 3.81 & 4.78 \\
 & 8        & 3.46 & 2.05 & 3.58 & 4.59 \\
\midrule

\multirow{2}{*}{4}
 & $\infty$ & 4.15 & 2.31 & 3.98 & 4.74 \\
 & 8        & 3.86 & 2.26 & 3.70 & 4.65 \\
\midrule

\multirow{2}{*}{8}
 & $\infty$ & 4.30 & 2.36 & 4.96 & 4.76 \\
 & 8        & 3.95 & 2.30 & 4.56 & 4.64 \\
\bottomrule
\end{tabular}
\end{adjustbox}
\end{table}

\begin{table}[ht]
\centering
\caption{MAPE results for GNN-only in $D_2$ for different $T_{inGNN}$ and $F_T$, reorganized for single-column layout.}
\label{tab:gnn_only_d2_2}
\begin{adjustbox}{max width=\linewidth}
\begin{tabular}{c c | c c c | c c c}
\toprule
\multicolumn{2}{c|}{} & \multicolumn{3}{c|}{Cont} & \multicolumn{3}{c}{Rand} \\
\textbf{$T_{inGNN}$} & \textbf{$F_T$} & MILC & LAMMPS & NN & MILC & LAMMPS & NN \\
\hline
\multirow{2}{*}{2} & $\infty$
  & 4.26 & 5.46 & 3.60
  & 3.93 & 4.88 & 3.51 \\
 & 8
  & 3.85 & 4.91 & 3.64
  & 3.69 & 4.68 & 3.44 \\
\hline
\multirow{2}{*}{4} & $\infty$
  & 4.63 & 5.43 & 4.14
  & 4.11 & 4.68 & 3.88 \\
 & 8
  & 4.41 & 5.57 & 3.80
  & 3.82 & 4.59 & 3.84 \\
\hline
\multirow{2}{*}{8} & $\infty$
  & 4.56 & 5.28 & 4.29
  & 5.24 & 4.80 & 4.01 \\
 & 8
  & 4.27 & 5.24 & 4.37
  & 4.82 & 4.68 & 3.89 \\
\bottomrule
\end{tabular}
\end{adjustbox}
\end{table}

\begin{table}[ht]
\centering
\caption{MAPE results for LLM-only for different $T_{inLLM}$ for the simulation in $D_1$.}
\label{tab:timellm_only_d1}
\begin{adjustbox}{max width=\linewidth}
\begin{tabular}{c|cc|cc}
\toprule
\multirow{2}{*}{}
& \multicolumn{2}{c|}{Cont} 
& \multicolumn{2}{c}{Rand} \\
& \rotatebox{0}{MILC} 
& \rotatebox{0}{LAMMPS} 
& \rotatebox{0}{MILC} 
& \rotatebox{0}{LAMMPS} \\
\hline
$T_{inLLM} = 2$ & 4.23 & 2.77 & 4.24 & 2.77 \\
$T_{inLLM} = 4$ & 4.36 & 1.98 & 4.33 & 1.90 \\
$T_{inLLM} = 8$ & 4.02 & 1.94 & 3.97 & 1.91 \\
\bottomrule
\end{tabular}
\end{adjustbox}
\vspace{-4mm}
\end{table}

\begin{table}[ht]
\centering
\caption{MAPE results for LLM-only in $D_2$ for different $T_{inLLM}$.}
\label{tab:timellm_only_d2}
\begin{adjustbox}{max width=\linewidth}
\begin{tabular}{c|ccc|ccc}
\toprule
\multirow{2}{*}{} 
& \multicolumn{3}{c|}{Cont} 
& \multicolumn{3}{c}{Rand} \\
& \rotatebox{0}{MILC}
& \rotatebox{0}{LAMMPS }
& \rotatebox{0}{NN}
& \rotatebox{0}{MILC}
& \rotatebox{0}{LAMMPS}
& \rotatebox{0}{NN} \\

\hline

$T_{inLLM} = 2$ & 4.29 & 2.75 & 3.51 & 4.32 & 2.68 & 3.51 \\

$T_{inLLM} = 4$ & 4.48 & 2.08 & 3.48 & 4.51 & 1.97 & 3.44 \\

$T_{inLLM} = 8$ & 4.09 & 1.93 & 3.36 & 4.13 & 1.94 & 3.22 \\

\bottomrule
\end{tabular}
\end{adjustbox}
\end{table}

\begin{table}[h]
\centering
\caption{Runtime prediction MAPE (\%) on D2-cont when replacing \ourmodel’s Time-LLM with Autoformer or DLinear.}
\label{tab:temporal-ablation-mape}
\begin{tabular}{l|ccc}
\toprule
Model Variant & MILC & LAMMPS  & NN \\
\midrule
\ourmodel (Time-LLM)  & \textbf{3.91} & 1.93 & \textbf{3.07} \\
GNN + Autoformer      & 4.22          & \textbf{1.84} & 3.68 \\
GNN + DLinear         & 4.39          & 1.95 & 4.11 \\
GNN + PatchTST        & 4.05          & 1.91 & 3.15 \\
\bottomrule
\end{tabular}
\end{table}

\subsection{Results on Inference Time}
Table~\ref{table:inference_time} shows the average inference times for all models, demonstrating that they are orders of magnitude faster than traditional PDES simulations. Inference times range from under 0.00001 seconds (LAST) to 0.5150 seconds (\ourmodel). In contrast, Table~\ref{table:sim_time} shows that PDES iteration times span from 8.09 to 243.48 seconds, depending on the application and job placement. Even the slowest surrogate model (\ourmodel) requires less than 1\% of the shortest PDES iteration time and under 0.2\% of the longest, making real-time use feasible. These results confirm the surrogate models are lightweight and suitable for integration into time-sensitive simulation workflows.

\textbf{Discussion.} 
A practical surrogate model must execute efficiently to benefit large-scale simulations. While traditional simulations can take hours or days to complete, \ourmodel performs inference in just 0.5150 seconds per step, making hybrid simulation both efficient and feasible.

\subsection{Ablation Study}
\label{sec:ablation}

We evaluate the contributions of the two main components of \ourmodel: the GNN encoder and the Time-LLM module.

First, the GNN-only variant, which removes the LLM component and relies solely on the GNN encoder and temporal transformer, shows significantly degraded performance on $D_1$ (Table~\ref{tab:gnn_only_d1}), indicating that LLM plays a critical role in capturing temporal patterns. Accuracy also improves when using a shorter online tuning window ($F_T = 8$).
Table \ref{tab:gnn_only_d2_2} shows the MAPE results for the GNN-only model on dataset $D_2$. This model removes the LLM component from \ourmodel. In this dataset, the results are consistent with our findings for dataset $D_1$. The GNN-only model performs worse than the full \ourmodel model. That implies that the LLM component is a critical component for \ourmodel.

Next, the LLM-only variant excludes the GNN and uses only temporal input to predict future iteration times based on a look-back window $T_{inLLM}$. This model also underperforms \ourmodel on $D_1$ (Table~\ref{tab:timellm_only_d1}), underscoring the importance of spatial features. Longer input windows generally lead to better accuracy by allowing the model to leverage more historical data. 
Table \ref{tab:timellm_only_d2} presents the MAPE results for the LLM-only model on dataset $D_2$. The trend is similar to those ones on dataset $D_1$. The LLM-only model relies solely on the LLM module to predict future iteration times and also performs worse than the full \ourmodel model as expected. This indicates that the GNN-based component is needed to capture spatial dependencies.

\subsection{Parameter or Hyper-parameter variations}
We analyze the impact of varying key hyperparameters to better understand their effect on prediciton accuracy. We observe only minor variations across hyperparameter settings, suggesting that SMART is robust and does not require extensive tuning to achieve strong performance.

\textit{Effect of Patch Length:} Figure \ref{fig:mape_patch_length} illustrates the results (MAPE) produced by \ourmodel when we vary the patch length of its LLM component. As observed, increasing the patch length slightly degrades the accuracy. This could be due to the lower granularity in capturing the temporal variations. 

\textit{Varying GCN Layers:} Figure \ref{fig:mape_gcn_layers} shows the impact of varying the number of GCN layers on the produced MAPE by \ourmodel. Increasing the number of layers beyond two does not lead to a significant accuracy gain. In fact, deeper GCNs may introduce over-smoothing effects that hinders performance. This also implies that a surrogate model only needs a local structure from the network topology to predict application run time accurately in the Dragonfly systems. 

\textit{Varying LLM hidden layers:} The effect of the variation of the number of hidden layers in the  LLM model is shown in Figure \ref{fig:mape_llm_layers}. The variation does not have significant effect on the MAPE.

\subsection{Temporal Model Comparisons}
\label{temporal-ablation}
We evaluate the role of \ourmodel’s Time-LLM by replacing it with three widely used time-series models: Autoformer, DLinear and PatchTST. All models used the same prompt-based input (T$_\text{in}$ = 8) and fixed GNN encoding (T$_\text{in,GNN}$ = 2). Results on the D2-cont dataset are reported in Table~\ref{tab:temporal-ablation-mape} using mean absolute percentage error (MAPE).

\ourmodel with Time-LLM outperforms both alternatives on MILC and NN workloads, while Autoformer achieves slightly better accuracy on LAMMPS. These results suggest that the LLM-based architecture in \ourmodel provides more consistent performance across diverse workload behaviors, especially those with bursty or nonstationary iteration patterns.

\section{Conclusion}
We presented \ourmodel, a domain-aware surrogate model designed to accelerate parallel discrete event simulation for Dragonfly interconnects. 
By integrating GNNs to model the hierarchical topology of Dragonfly networks and LLMs to capture temporal traffic dynamics, \ourmodel effectively learns both the spatial structure and temporal behavior intrinsic to Dragonfly systems.
Evaluated on 1,056-node Dragonfly simulations, \ourmodel consistently outperforms baseline methods in prediction accuracy and delivers sub-second inference, reducing simulation time from hours to seconds.

Although this study centers on Dragonfly systems, the modular architecture of \ourmodel allows for straightforward adaptation to other network topologies. For instance, applying \ourmodel to alternative architectures like Fat Tree requires only retraining with new graph-structured inputs, without altering the core model. 
In future work, we plan to extend the model's forecasting horizon and explore its generalization across network types. 

\section{Acknowledgment}
This work was supported in part by the U.S. Department of Energy under Contract No. DE-SC0024271 and by the U.S. National Science Foundation under Grants OAC-2402901 and CCF-2515009. The authors thank Kevin Brown and Rob Ross at Argonne National Laboratory and Chris Carothers at RPI for their helpful feedback and discussions.

\bibliography{bibliography}

\end{document}